\documentclass[
]{ceurart}

\usepackage{listings}
\usepackage{bm}

\begin{document}

\copyrightyear{2022}
\copyrightclause{Copyright for this paper by its authors.
  Use permitted under Creative Commons License Attribution 4.0
  International (CC BY 4.0).}

\conference{Ital-IA 2025: 5th National Conference on Artificial Intelligence, organized by CINI, June 23--24, 2025, Trieste, Italy}

\title{An Agentic AI for a New Paradigm in Business Process Development}


\author[1]{Mohammad Azarijafari}[%
orcid=0000-0002-3018-7000,
email=mohammad.azarijafari@unitn.it,
]
\cormark[1]
\fnmark[1]
\address[1]{Department of Industrial Engineering, University of Trento, Via Sommarive 9, 38123, Trento, Italy}

\author[1]{Luisa Mich}[%
orcid=0000-0002-0018-6883,
email=luisa.mich@unitn.it,
]
\fnmark[1]

\author[2]{Michele Missikoff}[%
orcid=0000-0002-7972-5201,
email=michele.missikoff@iasi.cnr.it,
]
\fnmark[1]
\address[2]{Istituto di Analisi dei Sistemi ed Informatica (IASI) “Antonio Ruberti”, National Research Council (CNR), Via dei Taurini 19, 00185, Rome, Italy}

\cortext[1]{Corresponding author.}
\fntext[1]{These authors contributed equally.}

\begin{abstract}
  Artificial Intelligence agents represent the next major revolution in the continuous technological evolution of industrial automation. In this paper, we introduce a new approach for business process design and development that leverages the capabilities of Agentic AI. Departing from the traditional task-based approach to business process design, we propose an agent-based method, where agents contribute to the achievement of business goals, identified by a set of business objects. When a single agent cannot fulfill a goal, we have a merge goal that can be achieved through the collaboration of multiple agents. The proposed model leads to a more modular and intelligent business process development by organizing it around goals, objects, and agents. As a result, this approach enables flexible and context-aware automation in dynamic industrial environments.
\end{abstract}


\maketitle

\section{Introduction}

The recent rapid advancement in digital environments has led businesses to increasingly look for intelligent, adaptable, and autonomous systems to enhance their operations. Traditional business process (BP) models, which are typically based on predefined task sequences and static rules, cannot meet the demands of dynamic markets and complex organizational ecosystems \cite{debenham2002multi}. These limitations have motivated a shift towards more flexible and context-aware approaches to BP design, development, and execution. Such approaches are now possible by the emergence of advanced AI technologies.

One of the most promising developments in this area is the advent of agentic AI, a class of artificial intelligence systems that operate through autonomous agents \cite{acharya2025agentic}. Agentic AI systems operate without the need for continuous human intervention and can independently make decisions, pursue goals, and adapt to changing contexts. These agents are capable of long-term planning and proactive behavior, which allows them to go beyond simple task execution and contribute meaningfully to dynamic workflows. Their autonomy is based on mechanisms such as language understanding, reasoning engines, memory modules, and reinforcement learning \cite{plaat2025agentic}.

This paper introduces an approach to business process development based on the capabilities of agentic AI, with a particular focus on systems powered by Large Language Models (LLMs) and Generative AI (GenAI). We propose a method in which BPs are not defined by fixed workflows but rather by business goals, information objects, and autonomous agents responsible for achieving them. This represents a shift from a task-based model to an agent-based goal-driven model, in which workflows emerge from agent interactions rather than being predesigned.

In the following sections, we first introduce the core concepts of this paradigm, then we outline our proposal using a running example to better clarify the approach, and finally, we formalize the proposed method for agent-based business process.

\section{Related Work}

Recent research highlights a significant movement toward integrating autonomous AI agents into business processes, driven by advancements in generative AI and multi-agent systems. Vu et al., through a review of three decades of research, emphasized the necessity for robust methodologies to manage agent autonomy and mitigate associated risks \cite{vu2025agentic}. Zhang et al. introduced EvoFlow, an evolutionary approach leveraging diverse LLMs to dynamically optimize agentic workflows \cite{zhang2025evoflow}. Similarly, Niu et al. proposed a new framework, and emphasized real-time adaptability and parallel execution using modularized agent architectures \cite{niu2025flow}.

To facilitate practical adoption, Jeong developed a multimodal multi-agent system via a No-Code platform, which reduced enterprise barriers to AI implementation \cite{jeong2025beyond}. Bousetouane further expanded on domain-specific agentic solutions, integrating reasoning, memory, and cognitive modules \cite{bousetouane2025agentic}. In parallel, Kandogan et al. presented a compound AI architecture for orchestrating agents, data streams, and workflows within enterprise contexts \cite{kandogan2025orchestrating}. Additionally, Tupe and Thube explored strategic API frameworks designed specifically for supporting agent-driven workflows in dynamic organizational environments \cite{tupe2025ai}.

Collectively, these studies highlight advancements in agent-based BP, yet a gap remains for goal-driven methods appropriate for dynamic business environments. To address this gap, this paper proposes an adaptable agentic AI approach tailored for real-time business scenarios.

\section{Agent-Based BP Automation}

The main assumption of our method is that a BP can be seen as a coordinated team of AI agents. In other word, the approach provides an agent-based view of a BP rather than the traditional task-oriented view. Essentially, we focus on the ‘what’, i.e., goals, objects, and capabilities, rather than the ‘how’, i.e., the tasks. In this respect, our agent-based approach is mainly a declarative one, and it is different from the traditional ones based on workflows, where the central elements of a BP are the tasks and a BP is modelled as a partially ordered set of tasks, typically represented by a process diagram.

The BP is built starting from three basic components: \textit{goals}, \textit{objects} and \textit{agents}. We start with the analysis of the BP to identify the business goals and sub-goals to be achieved. A \textbf{goal} is a desired state of affairs, represented by a set of business \textbf{objects}, i.e., information, in various possible forms (documents, messages, and database records, etc.) that will be generated by dedicated agents. An \textbf{agent} represents the active entity working for the BP to achieve its goals. Agents are defined by their goals, the manipulated objects, and the capabilities required to achieve the goals. An agent is activated when its trigger objects are ready, typically released by a preceding agent, or by a special object that causes a process to begin, referred to as the start object. In reaching its goal, the agent releases its final objects.

In essence, a BP is mainly represented by a partially order set of goals, resulting in a non-deterministic workflow, where nodes are goals and the arcs represent the agents responsible for achieving the former.

In an agent-based BP, there can be various alternatives to achieve a goal, i.e., different actions can be undertaken to reach the same goal. Sometimes such activities can be freely chosen, being equivalent, in other cases the choice may depend on the circumstances or existing constraints. The agent will be able to analyse the context and make the most convenient choice.

As anticipated, reaching a goal requires the creation of one or more objects in the form of a business document, a set of connected documents, a message, a data record in a database, or any other business information required during the execution of the BP. If an object is a physical one, we require that its digital image is created in parallel.

Precedence. In the context of business process modeling, precedence is a binary relation between two goals that imposes an order of achievement. This relation captures dependency constraints over goals that imposes the activation sequence of agents within a process. We do not need to explicitly provide such relations; they are inductively derived analysing the trigger objects of agents.

Objects are the passive entities manipulated by the agents. Objects can trigger the activation of an agent or represent a resource needed by an agent during its operations. Reaching its goal, an agent releases its final objects. There are two special sets of objects: one necessary for the process to start and one generated when the process terminates.

Capabilities refer to the operational skills and abilities necessary for an agent to operate effectively in accomplishing its goal. Such capabilities refer to the \textit{CRUDA} operations, where \textit{CRUDA} is an acronym proposed in the database theory and refers to the following operations: \textit{Create}, \textit{Read}, \textit{Update}, and \textit{Delete} operations, to which we add the \textit{Archive} operation that is particularly relevant in the business domain, where documents need to be preserved for future inspection, if required.

An agent definition includes the pursued goal, the \textit{capabilities} necessary to achieve the goal, the \textit{trigger objects} necessary to start and the \textit{final objects} released at the end of its operations. Finally, the \textit{resource objects} that the agent needs throughout its lifecycle for its operations.

The trigger objects have been released by another agent or provided at the beginning of the process (called \textit{start objects}). The trigger objects represent the input necessary for the agent to initiate its execution. Furthermore, the agent definition requires the specification of the output objects generated when it reaches its goal.

\begin{figure}[h!]
    \centering
    \includegraphics[width=0.8\linewidth]{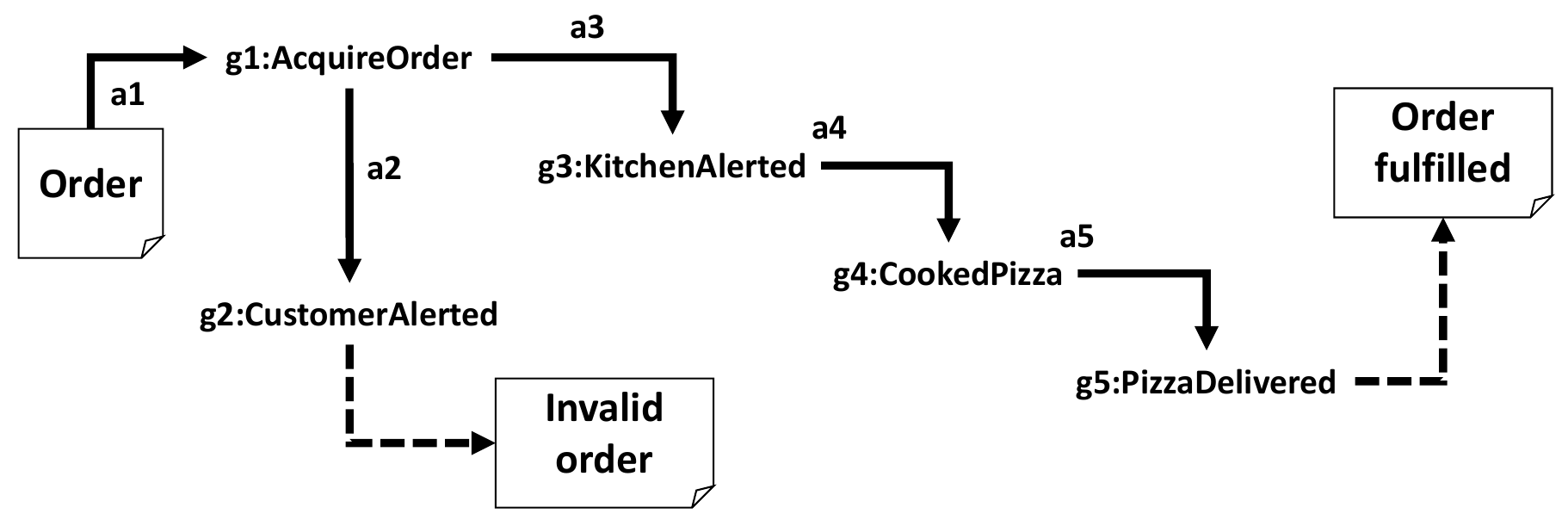}
    \caption{Agent-based workflow for the pizza delivery process.}
    \label{fig:pizza_delivery_process}
\end{figure}

Below we provide a diagram that describes a simple example of a home delivery pizza shop (Figure \ref{fig:pizza_delivery_process}), and in particular the process that starts with the customer order and terminates successfully with the creation of ‘fulfilled order’ object. As anticipated, the nodes represent the goals and the arcs are labelled with agents ID. In this simple example there is only one split node, in case the submitted order is for some reasons incorrect. The CookedPizza is a physical object that in the system is represented by a database record. The solid arcs represent the agent operations, the dotted arcs imply the release of a document, as a final operation of the agent.

Below we have a simplified tabular representation of the knowledge base that represents the diagram. Note that for sake of compactness we omitted the resource objects (Table \ref{tab:pizza_workflow_specifications}).

\begin{table*}[h!]
    \caption{Agent-based workflow specifications for the pizza delivery process.}
    \label{tab:pizza_workflow_specifications}
    \begin{tabular}{cllll}
        \toprule
        Agent ID & Competences       & Trigger objects & Final objects   & Goal            \\
        \midrule
        a1       & Get\&CheckOrder   & order           & checkedOrder    & AcquireOrder    \\
        a2       & InformingCustomer & checkedOrder/KO & customerNotice  & CustomerAlerted \\
        a3       & InformingKitchen  & checkedOrder/OK & pizzaSchedule   & KitchenAlerted  \\
        a4       & CookPizza         & pizzaSchedule   & pizzaDone       & CookedPizza     \\
        a5       & Delivering        & pizzaDone       & fullfilledOrder & PizzaDelivered  \\ 
        \bottomrule
    \end{tabular}
\end{table*}

Please note that with trigger objects it is possible to derive which are the agents that need to be executed. In case of a merge goal, the objects associated with the goal, $\bm{OG}$, will be obtained by the union of al the incoming agents: $\bm{OG_g} = U_i \bm{OF_i}$, where \textit{i} is the index of all incoming agents.

\section{Formal Representation of Agent-Based BP}

The reported example depicts a simplified case, where no parallel paths are presented. In general, we have \textit{split goals}, where more than one agent is triggered. Symmetrically, there have \textit{merge goals} that are satisfied by the union of the released objects of two or more agents.

An agent is a 6-tuple:

\begin{center}
    $\bm{Agent} = \bm{(aID, C_a, OT_a, OR_a, OF_a, g_a)}$ 
\end{center}

\begin{itemize}
    \item $\bm{aID}$ is the agent identifier.
    \item $\bm{C_a}$ is the set of capabilities, essentially CRUDA operations.
    \item $\bm{OT_a}$: Triggering objects necessary for the agent to wake up and start its operations.
    \item $\bm{OR_a}$: Resource objects, tackled during the agent’s operations.
    \item $\bm{OF_a}$ is the set of objects released by the agent at the end of its operations when it reaches its goal $\bm{g_a}$.
    \item $\bm{g_a}$ is the goal of agent $\bm{a}$.
\end{itemize}

An agent $\bm{a}$ functionally determines the pair $\bm{(gID, g_a)}$, where the first element is the goal that triggers the agent, and the second is the agent’s final goal:

\begin{center}
    $\bm{aID} \rightarrow \bm{(gID_a, g_a)}$
\end{center}

In general, the set of objects in the scope of an agent $\bm{a}$ is 

\begin{center}
    $\bm{O_a} = \bm{OT_a} \cup \bm{OR_a} \cup \bm{OF_a}$
\end{center}

Note that an agent starts its operations when its $\bm{OT_a}$ is ready. This typically takes place when the previous agent(s) has/have reached its/their goal(s).

In formal terms, a goal is represented by the following triple:

\begin{center}
    $\bm{g} = \bm{(gID, O_g, A_g)}$
\end{center}

where:

\begin{itemize}
    \item $\bm{gID}$ is the goal identifier.
    \item $\bm{O_g}$ is the set of objects that characterises the goal. $\bm{O_g}$ can be built by more than one agent, in case of a merge goal, then it is the set union of the final objects of each incoming agent.
    \item Ag is the set of agents triggered by $\bm{O_g}$. If the set A is not a singleton, then the goal is referred to as a split goal. It means that more than one agent may be triggered when the goal is achieved. We have three types of splits: $AND$, $OR$, $XOR$.
\end{itemize}

In the case of a split goal, the associated set of objects is able to trigger more than one agent. Depending on the trigger objects of the agents, the activation can be parallel (all the agent will start simultaneously: $AND$ condition) or in a disjunctive ($OR$, $XOR$).

A merge goal, is a node of the graph with more than one incident agent. The goal will be fully fulfilled when all the agents terminate ($AND$ merge) or at least one agent reaches its end ($OR$, $XOR$ merge).

It is useful to introduce the precedence relation, $\bm{pre}$, that allow us to draw a goal sequencing diagram as reported in the example of Figure \ref{fig:pizza_delivery_process}. Given two goals, say $\bm{gx}$ and $\bm{g_y}$, we say that $\bm{gx}$ strictly precedes $\bm{g_y}$ if there are agents in $\bm{A_x}$ such that, once triggered, directly contribute to the achievement of $\bm{Oy}$. Then, we have: $\bm{pre(g_x, g_y)}$.

\section{Agent-Based Goal-Driven BP}

Considering the previous definitions of agents and goals, we can define an \textit{Agent-based BP (ABP)}, as a 6-tuple:

\begin{center}
    $\bm{ABP} = \bm{(OS, OE, OR, G, C, A)}$
\end{center}

where:

\begin{itemize}
    \item $\bm{OS}$ is the set of objects that triggers the first agent(s) of the BP.
    \item $\bm{OE}$ is the final set of objects released by the BP that corresponds to the achievement of the last agents.
    \item $\bm{OR}$ is the set of all resource objects relevant to the BP.
    \item $\bm{G}$ is a set of goals.
    \item $\bm{C}$ is the set of all the capabilities necessary to carry out the BP.
    \item $\bm{A}$ is the set of actors required for the execution of the BP. Intuitively, you can see a set of agents inducing a covering over $\bm{C}$.
\end{itemize}

We are aware that the definition of BP is redundant, but such a redundancy, in the analysis phase, is helpful to achieve a number of check and to verify the correctness of the BP specification. In particular, the precedence relation is used to check if for the agents of a BP there is a correspondence between the trigger objects and the objects associated to the goals. For instance, trigger objects (except those in $\bm{OS}$) that do not belong to any goal raise a red flag, since there is an agent that will never be able to wake up. Symmetrically, an object in $\bm{o_x}$ that does not appear in any $\bm{OT_y}$ is a redundant object. Furthermore, the $\bm{pre}$ relation imposes an ordering constraint that must be respected during the execution of the business process.

\section{Conclusion}

Traditional task-based approaches to business process development often fall short when it comes to meeting the demands of today’s dynamic and complex organizational environments. They typically lack the flexibility and autonomy needed to adapt in real time. To address these challenges, we introduce a new approach grounded in agentic AI, where autonomous agents work together to carry out business processes in dynamic settings. Our model shifts the focus from fixed task flows to business goals, information objects, and intelligent agents. This shift enables a more modular and goal-driven design, and making it easier to implement flexible and adaptive automation in real-time business conditions.

However, with great potential comes significant challenges. The autonomy of agentic systems raises crucial questions about safety, ethics, accountability, and control. How do we ensure these systems align with human intentions? What safeguards are needed to prevent unintended consequences, especially when agents operate at scale or across interconnected domains? Furthermore, the shift from passive tools to active agents requires rethinking human-AI collaboration, trust, and responsibility.

Another key concern is governance. Agentic AI amplifies both the capabilities and the risks of existing AI technologies. Ensuring transparency in how agents make decisions, maintaining human oversight, and creating mechanisms for audit and correction are essential steps toward responsible deployment. As such, the development of agentic AI must go hand in hand with robust regulatory frameworks and interdisciplinary dialogue.

In summary, agentic AI represents a new frontier in artificial intelligence, one that brings machines closer to behaving as active participants in our digital ecosystems. While the promise of these technologies is important, their implementation requires thoughtful design, careful oversight, and a deep understanding of the mechanisms governing machine agency.

\bibliography{sample-ceur}

\begin{thebibliography}{10}
\expandafter\ifx\csname natexlab\endcsname\relax\def\natexlab#1{#1}\fi
\providecommand{\url}[1]{\texttt{#1}}
\providecommand{\href}[2]{#2}
\providecommand{\path}[1]{#1}
\providecommand{\DOIprefix}{doi:}
\providecommand{\ArXivprefix}{arXiv:}
\providecommand{\URLprefix}{URL: }
\providecommand{\Pubmedprefix}{pmid:}
\providecommand{\doi}[1]{\href{http://dx.doi.org/#1}{\path{#1}}}
\providecommand{\Pubmed}[1]{\href{pmid:#1}{\path{#1}}}
\providecommand{\bibinfo}[2]{#2}
\ifx\xfnm\relax \def\xfnm[#1]{\unskip,\space#1}\fi
\bibitem[{Debenham(2002)}]{debenham2002multi}
\bibinfo{author}{J.~Debenham},
\newblock \bibinfo{title}{A multi-agent architecture for business process management adapts to unreliable performance},
\newblock in: \bibinfo{booktitle}{Adaptive Computing in Design and Manufacture V}, \bibinfo{publisher}{Springer}, \bibinfo{year}{2002}, pp. \bibinfo{pages}{369--380}. \DOIprefix\doi{10.1007/978-0-85729-345-9_31}.
\bibitem[{Acharya et~al.(2025)Acharya, Kuppan, and Divya}]{acharya2025agentic}
\bibinfo{author}{D.~B. Acharya}, \bibinfo{author}{K.~Kuppan}, \bibinfo{author}{B.~Divya},
\newblock \bibinfo{title}{Agentic ai: Autonomous intelligence for complex goals--a comprehensive survey},
\newblock \bibinfo{journal}{IEEE Access}  (\bibinfo{year}{2025}). \DOIprefix\doi{10.1109/ACCESS.2025.3532853}.
\bibitem[{Plaat et~al.(2025)Plaat, van Duijn, van Stein, Preuss, van~der Putten, and Batenburg}]{plaat2025agentic}
\bibinfo{author}{A.~Plaat}, \bibinfo{author}{M.~van Duijn}, \bibinfo{author}{N.~van Stein}, \bibinfo{author}{M.~Preuss}, \bibinfo{author}{P.~van~der Putten}, \bibinfo{author}{K.~J. Batenburg},
\newblock \bibinfo{title}{Agentic large language models, a survey},
\newblock \bibinfo{journal}{arXiv preprint arXiv:2503.23037}  (\bibinfo{year}{2025}). \DOIprefix\doi{10.48550/arXiv.2503.23037}.
\bibitem[{Vu et~al.(2025)Vu, Klievtsova, Leopold, Rinderle-Ma, and Kampik}]{vu2025agentic}
\bibinfo{author}{H.~Vu}, \bibinfo{author}{N.~Klievtsova}, \bibinfo{author}{H.~Leopold}, \bibinfo{author}{S.~Rinderle-Ma}, \bibinfo{author}{T.~Kampik},
\newblock \bibinfo{title}{Agentic business process management: The past 30 years and practitioners' future perspectives},
\newblock \bibinfo{journal}{arXiv preprint arXiv:2504.03693}  (\bibinfo{year}{2025}).
\bibitem[{Zhang et~al.(2025)Zhang, Chen, Wan, Chang, Cheng, Wang, Hu, and Bai}]{zhang2025evoflow}
\bibinfo{author}{G.~Zhang}, \bibinfo{author}{K.~Chen}, \bibinfo{author}{G.~Wan}, \bibinfo{author}{H.~Chang}, \bibinfo{author}{H.~Cheng}, \bibinfo{author}{K.~Wang}, \bibinfo{author}{S.~Hu}, \bibinfo{author}{L.~Bai},
\newblock \bibinfo{title}{Evoflow: Evolving diverse agentic workflows on the fly},
\newblock \bibinfo{journal}{arXiv preprint arXiv:2502.07373}  (\bibinfo{year}{2025}). \DOIprefix\doi{10.48550/arXiv.2502.07373}.
\bibitem[{Niu et~al.(2025)Niu, Song, Lian, Shen, Yao, Zhang, and Liu}]{niu2025flow}
\bibinfo{author}{B.~Niu}, \bibinfo{author}{Y.~Song}, \bibinfo{author}{K.~Lian}, \bibinfo{author}{Y.~Shen}, \bibinfo{author}{Y.~Yao}, \bibinfo{author}{K.~Zhang}, \bibinfo{author}{T.~Liu},
\newblock \bibinfo{title}{Flow: A modular approach to automated agentic workflow generation},
\newblock \bibinfo{journal}{arXiv preprint arXiv:2501.07834}  (\bibinfo{year}{2025}). \DOIprefix\doi{10.48550/arXiv.2501.07834}.
\bibitem[{Jeong(2025)}]{jeong2025beyond}
\bibinfo{author}{C.~Jeong},
\newblock \bibinfo{title}{Beyond text: Implementing multimodal large language model-powered multi-agent systems using a no-code platform},
\newblock \bibinfo{journal}{arXiv preprint arXiv:2501.00750}  (\bibinfo{year}{2025}). \DOIprefix\doi{10.13088/jiis.2025.31.1.191}.
\bibitem[{Bousetouane(2025)}]{bousetouane2025agentic}
\bibinfo{author}{F.~Bousetouane},
\newblock \bibinfo{title}{Agentic systems: A guide to transforming industries with vertical ai agents},
\newblock \bibinfo{journal}{arXiv preprint arXiv:2501.00881}  (\bibinfo{year}{2025}). \DOIprefix\doi{10.48550/arXiv.2501.00881}.
\bibitem[{Kandogan et~al.(2025)Kandogan, Bhutani, Zhang, Chen, Gurajada, and Hruschka}]{kandogan2025orchestrating}
\bibinfo{author}{E.~Kandogan}, \bibinfo{author}{N.~Bhutani}, \bibinfo{author}{D.~Zhang}, \bibinfo{author}{R.~L. Chen}, \bibinfo{author}{S.~Gurajada}, \bibinfo{author}{E.~Hruschka},
\newblock \bibinfo{title}{Orchestrating agents and data for enterprise: A blueprint architecture for compound ai},
\newblock \bibinfo{journal}{arXiv preprint arXiv:2504.08148}  (\bibinfo{year}{2025}). \DOIprefix\doi{10.48550/arXiv.2504.08148}.
\bibitem[{Tupe and Thube(2025)}]{tupe2025ai}
\bibinfo{author}{V.~Tupe}, \bibinfo{author}{S.~Thube},
\newblock \bibinfo{title}{Ai agentic workflows and enterprise apis: Adapting api architectures for the age of ai agents},
\newblock \bibinfo{journal}{arXiv preprint arXiv:2502.17443}  (\bibinfo{year}{2025}). \DOIprefix\doi{10.48550/arXiv.2502.17443}.

\end{thebibliography}




\end{document}